\documentclass[10pt,twocolumn,letterpaper]{article}

\usepackage[pagenumbers]{cvpr} 

\usepackage[accsupp]{axessibility}
\usepackage{graphicx}
\usepackage{amsmath}
\usepackage{amssymb}
\usepackage{bbm}
\usepackage{booktabs}
\usepackage{xfrac}
\usepackage{multirow}
\usepackage{pifont}
\usepackage{colortbl}

\usepackage{stfloats}
\usepackage{float}
\usepackage{xcolor}
\usepackage{makecell}
\usepackage{wrapfig,lipsum}

\usepackage[pagebackref,breaklinks,colorlinks]{hyperref}

\usepackage[capitalize]{cleveref}
\crefname{section}{Sec.}{Secs.}
\Crefname{section}{Section}{Sections}
\Crefname{table}{Table}{Tables}
\crefname{table}{Tab.}{Tabs.}
\crefname{equation}{Eq.}{Eqs.}

\def\cvprPaperID{5057} 
\def\confName{CVPR}
\def\confYear{2022}

\usepackage{overpic}
\usepackage{enumitem} 
\usepackage{overpic} 
\usepackage{color}

\definecolor{turquoise}{cmyk}{0.65,0,0.1,0.3}
\definecolor{purple}{rgb}{0.65,0,0.65}
\definecolor{dark_green}{rgb}{0, 0.5, 0}
\definecolor{orange}{rgb}{0.8, 0.6, 0.2}
\definecolor{red}{rgb}{0.8, 0.2, 0.2}
\definecolor{darkred}{rgb}{0.6, 0.1, 0.05}
\definecolor{blueish}{rgb}{0.0, 0.3, .6}
\definecolor{light_gray}{rgb}{0.7, 0.7, .7}
\definecolor{pink}{rgb}{1, 0, 1}
\definecolor{greyblue}{rgb}{0.25, 0.25, 1}

\newcommand{\Table}[1]{Table~\ref{tab:#1}}

\usepackage{blindtext}

\renewcommand{\paragraph}[1]{\vspace{1em}\noindent\textbf{#1}.}
\begin{document}
\title{Cannot See the Forest for the Trees: \\ Aggregating Multiple Viewpoints to Better Classify Objects in Videos}



\author{
Sukjun Hwang$^1${\qquad}Miran Heo$^1${\qquad}Seoung Wug Oh$^2${\qquad}Seon Joo Kim$^1$
\vspace{1mm}\\$^1$Yonsei University\qquad$^2$Adobe Research
\\\texttt{\normalsize{\{sj.hwang, miran, seonjookim\}@yonsei.ac.kr\quad seoh@adobe.com}}
\vspace{-2mm}
}
\maketitle
\begin{abstract}
Recently, both long-tailed recognition and object tracking have made great advances individually.
TAO benchmark presented a mixture of the two, long-tailed object tracking, in order to further reflect the aspect of the real-world.
To date, existing solutions have adopted detectors showing robustness in long-tailed distributions, which derive per-frame results.
Then, they used tracking algorithms that combine the temporally independent detections to finalize tracklets.
However, as the approaches did not take temporal changes in scenes into account, inconsistent classification results in videos led to low overall performance.
In this paper, we present a set classifier that improves accuracy of classifying tracklets by aggregating information from multiple viewpoints contained in a tracklet.
To cope with sparse annotations in videos, we further propose augmentation of tracklets that can maximize data efficiency.
The set classifier is plug-and-playable to existing object trackers, and highly improves the performance of long-tailed object tracking.
By simply attaching our method to QDTrack on top of ResNet-101, we achieve the new state-of-the-art, 19.9\% and 15.7\% TrackAP$_\textit{50}$ on TAO validation and test sets, respectively.
Our code is available at this link\footnote{\texttt{\textcolor{blue}{\href{https://github.com/sukjunhwang/set\_classifier}{https://github.com/sukjunhwang/set\_classifier}}}}.
\vspace{-3mm}
\end{abstract}
\section{Introduction}
\label{sec:intro}
Object tracking is a long standing problem in computer vision as it plays a key role in surveillance and self-driving applications. 
There are numerous datasets and benchmarks for tracking~\cite{KITTI, MOT, MOTS, PoseTrack, VOT} and also a long list of tracking algorithms~\cite{SORT, CenterTrack, TransTrack, NeuralSolver, UniTrack}.
As with many other computer vision tasks, the performance of tracking algorithms has also taken a leap with deep learning.

Even with the great progress in object tracking, the performance of state-of-the-art trackers starts to degrade in the real-world scenarios with a large vocabulary of objects~\cite{TAO}.
This is because most tracking benchmarks include only a small set of objects such as pedestrian, vehicles, and animals, for targeting specific applications like autonomous driving.
To deploy the trackers in the real-world in a general environment, it is essential for the trackers to be able to deal with a much larger set of objects as in the image detection problem~\cite{LVIS}.
For this purpose, a new benchmark called TAO~\cite{TAO} for tracking any object has been recently introduced. 
This dataset contains over 800 categories, an order of magnitude more than previous tracking benchmarks. 

\begin{figure}[t]
\begin{center}
\includegraphics[width=\linewidth]{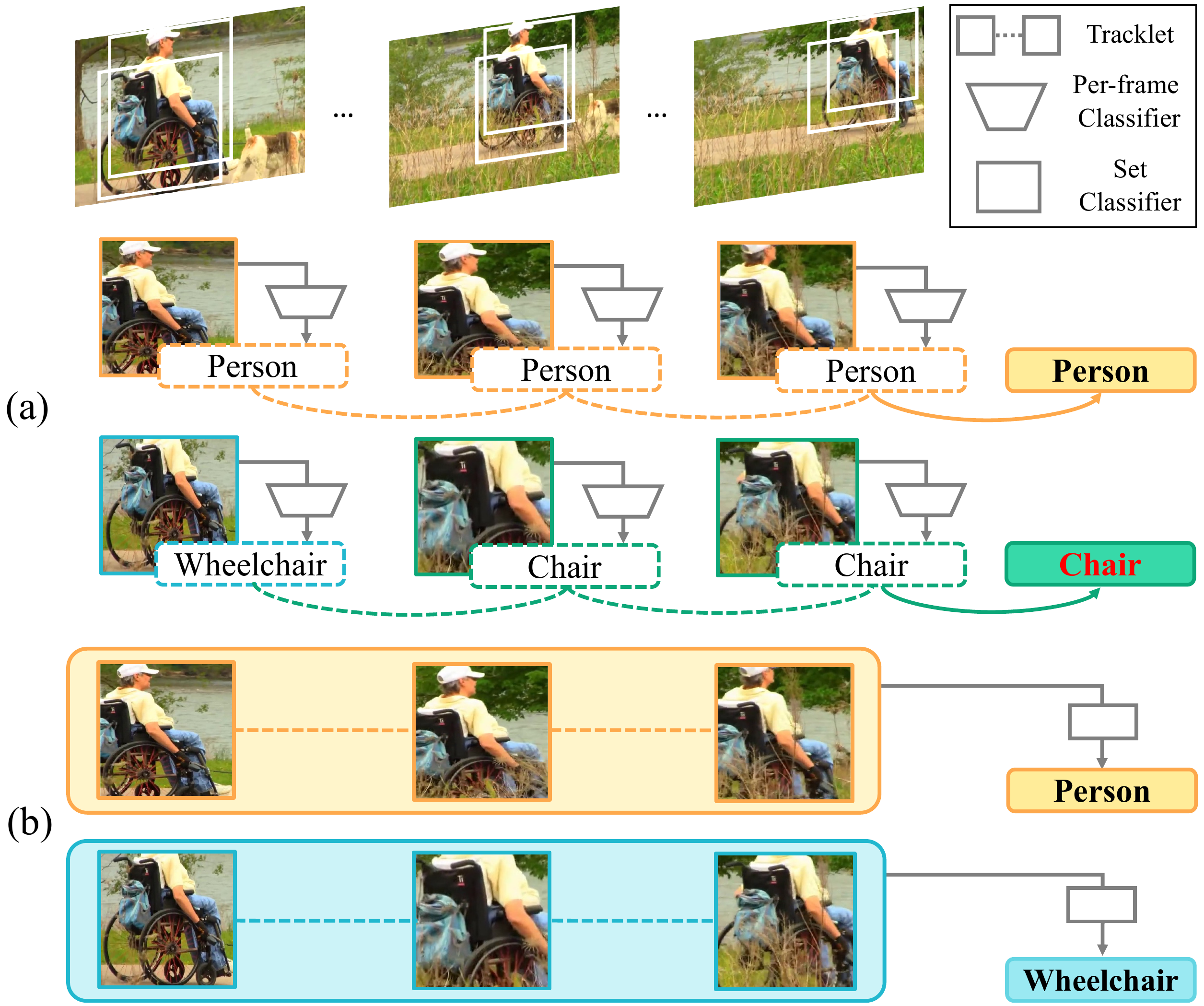}
\end{center}
\vspace{-6mm}
\caption{
(a) \textit{Per-frame classifier} receiving an instantaneous scene struggles on tail categories (e.g., wheelchair). On the other hand, (b) our proposed \textit{set classifier} shows robustness on tail categories by 
aggregating multiple viewpoints of a tracklet, taking the whole spatio-temporal feature into account.
}
\label{fig:teaser}
\vspace{-5mm}
\end{figure}

In \cite{TAO}, it was shown that most up-to-date trackers do not adapt well with increased number of object vocabulary.
While tracking algorithms have focused on accurately finding object boxes and tracking them, less attention has been paid on the classification of objects, primarily due to a small set of vocabulary. 
As the object category grows to a realistic size, classification becomes crucial for the overall tracking performance. 
After a thorough analysis, it was suggested in \cite{TAO} that \textit{``large-vocabulary tracking requires jointly improving tracking and classification accuracy''}.

In this paper, we show that aggregating multiple viewpoints of a tracklet is the key to classifying the large vocabulary in videos.
A tracklet refers to a set of boxes in different frames that share the same identity.
Although appearance of objects in tracklets may go through great changes, existing methods~\cite{SORT, TAO, QDTrack} determine the category of a tracklet from the collection of per-frame classification results as shown in \cref{fig:teaser} (a).
Since changes in scenes from temporal variations are not considered, they are vulnerable to appearance changes including motion blur or occlusion.
More importantly such cases bring critical deterioration of performance in tail classes.
Specifically, detectors trained on imbalanced data are more confident in frequent classes, and  such cases bring critical deterioration of performance in tail classes.

To this end, we propose a \textit{set classifier} that takes the spatio-temporal features of a whole tracklet into account (\cref{fig:teaser} (b)).
With this design, the set classifier is supplied with sufficient information to determine a category from the large vocabulary.
Therefore, the set classifier gains robustness against temporal shifts and the ability to avoid a collapse of final predictions from transient failures, leading to noticeable improvements of accuracy in the tail.

To fully supervise the set classifier to obtain the ability of exploiting spatio-temporal information, the module gets trained with video data.
In contrast to existing methods that can only classify the large vocabulary using frame-wise detections~\cite{TAO, SORT, QDTrack}, the essence of the set classifier is the ability to evaluate a whole tracklet by aggregating information from multiple sources.
The structural design of the set classifier is simple yet powerful; it is a stack of a few transformer layers~\cite{Transformer}.
Receiving multiple regional features~\cite{MaskRCNN} corresponding to the predicted boxes that compose a tracklet, the set classifier attends to relevant information that are necessary for the classification of the large vocabulary.

However, due to immense efforts required to annotate a video~\cite{TAO}, the annotation budget is insufficient to give the supervision of classifying tracklets under the complicated long-tailed scenarios.
As a solution to this dilemma, we present augmentation methods of generating tracklets that have video characteristics: a variety of viewpoints of an object.
Specifically, tracklets are dynamically generated utilizing regional proposals~\cite{FasterRCNN} from multiple source videos and images.
To further make the most out of the limited number of annotations, the augmented tracklets can be composed of multiple identities, and we introduce a training procedure for the set classifier using such tracklets.
With our methods, an enormous number of tracklet samples composed of rare classes can be obtained, and the set classifier gains the ability to successfully distinguish the large vocabulary in videos.

Adoption of our set classifier results in high performance improvement in the long-tailed tracking.
With the plug-and-playable design, we show experimental results on top of recently proposed QDTrack~\cite{QDTrack}, and achieve new state-of-the-art on the challenging TAO~\cite{TAO} benchmark: 19.9\% and 15.7\% TrackAP$_{50}$ on validation and test sets, respectively.
Furthermore, taking the same approach, we also achieve a competitive result of 37.7\% AP on the video instance segmentation dataset, YouTube-VIS 2019~\cite{MaskTrackRCNN}.

Our work can be summarized as follows:
\begin{itemize}[leftmargin=*]
\setlength\itemsep{-.3em}
\item We propose the set classifier which classifies a tracklet as a whole by aggregating information from multiple viewpoints.
\item We introduce augmentation methods that can generate augmented tracklets of near infinite diversities -- unlimited number of tracklets of tail classes can be obtained.
\item We propose a new training procedure that facilitates the supervision of the set classifier using the augmented tracklets.
Moreover, we suggest auxiliary losses that bring further improvements in accuracy.
\item We achieve new state-of-the-art on TAO, and also show the effectiveness of our method on YouTube-VIS 2019.
\end{itemize}


\section{Related works}
\label{sec:related}

\begin{figure*}[t]
\begin{center}
\includegraphics[width=\linewidth]{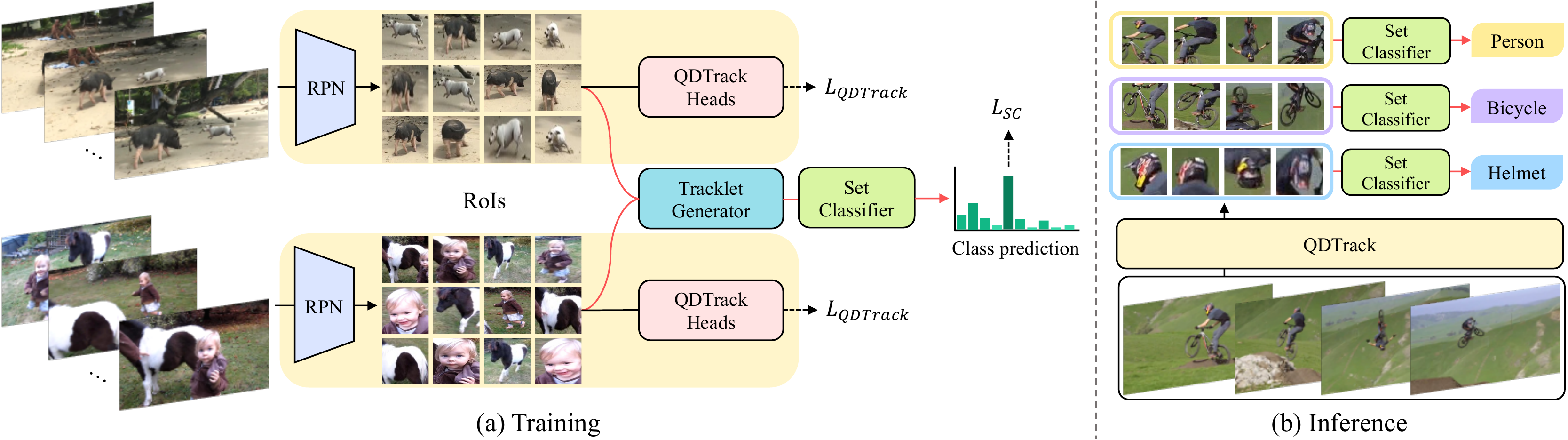}
\end{center}
\vspace{-5mm}
\caption{
Overview of our framework.
The set classifier is plug-and-playable to existing object trackers.
(a) During training, the set classifier gets trained with QDTrack simultaneously by receiving tracklets from the tracklet generator.
(b) For the inference, the object tracker first predicts tracklets.
Then, the set classifier takes a tracklet as a whole by fetching RoI tokens that correspond to boxes of the tracklet.
Finally, the proposed module comes out with re-classified predictions for each tracklet.
}
\vspace{-3mm}
\label{fig:sampling}
\end{figure*}

\paragraph{Long-Tailed Recognition}
While considerable progress in visual perception has been made on class-balanced datasets~\cite{ImageNet, COCO, Cityscapes}, naïve migration of cutting-edge models to long-tailed datasets~\cite{ImageNetLT, LVIS, LongtailOpenworld} shows severe failures.
With growing interests in inherent problems of long-tailed distributions, there have been noticeable explorations to the prevention of minor classes getting dominated by major classes.
Such solutions can be largely categorized into three: re-balancing distributions, re-weighting losses, and decoupled training.
Re-balancing methods~\cite{LVIS, LST, kang2019decoupling, chang2021image} tackle the sparsity of annotations in tail classes, which is the fundamental cause of long-tailed distributions.
The methods have shown that adequately sampling minor classes with higher rates can supplement the huge imbalance, showing improvements in the tail.
Instead of resolving the imbalance by weighting the sampling rate, the problem in long-tailed distributions can also be alleviated by differently weighting class-wise losses\cite{EQL, EQLv2, ClassBalancedLoss, Seesaw}.
The weights for different classes are determined with considerations to the number of annotations, mostly putting emphasis on tail classes.
BAGS~\cite{BAGS} and SimCal~\cite{SimCal} introduced attaching additional heads that are specialized in classifying tail classes, and combining the results from various heads during the inference.
Having categories grouped with a similar number of annotations, BAGS assigns different heads to each group, which prevents minor classes from getting dominated by major classes.
Similarly, the additional head of SimCal avoids the domination by receiving class-balanced proposals.

\paragraph{Understanding Objects in Videos}
From the huge success in images, there have been many tasks that shift the focus onto videos~\cite{MOTS, DAVIS, VOT, MaskTrackRCNN, MOT, PoseTrack}.
The fundamental characteristic of videos is that they are composites of a number of consecutive frames.
Therefore, well utilizing information and predictions from multiple frames has become the de facto factor in improving performance.

To exploit spatial proximity between consecutive frames, many approaches that target tracking tasks associate objects by fusing the motion prior with additional algorithms; optical flow~\cite{FlowTrack, HumanPoseOpticalFlow}, displacement regression~\cite{D&T, CenterTrack}, and Kalman Filter~\cite{KalmanFilter, SORT}.
However, relying heavily on the motion priors shows vulnerability to scenes with low frame rates or large camera motions.
To overcome those problems and handle reappearances, use of implicit features that represent objects can be considered~\cite{MaskTrackRCNN, RetinaTrack, QDTrack, GCT, NeuralSolver, TransTrack, TrackFormer}.

It has been shown that utilizing information from multiple frames is beneficial not only for tracking, but also for improving the quality of detection and segmentation in videos~\cite{STM, PCAN, MaskProp, VisTR, IFC}.
For example, in the video instance segmentation task~\cite{MaskTrackRCNN}, per-clip methods~\cite{PCAN, MaskProp, VisTR, IFC} generally show higher segmentation accuracy than approaches using only single frame information~\cite{MaskTrackRCNN, SipMask, CrossVIS, SGNet}.

\paragraph{Videos with Large Vocabulary}
As mentioned earlier, most previous video-related studies did not pay much attention to improving the classification performance.
Rather, major improvement came from tracking~\cite{MOT, VOT, PoseTrack}, detection~\cite{ImageNet}, and segmentation~\cite{MOTS, MaskTrackRCNN, DAVIS, OVIS}.
The primary reason for this trend is that most video related benchmarks include only a few major object classes.

Recently proposed benchmark TAO~\cite{TAO} is a video dataset with large vocabulary.
Since long-tailed scenarios were usually dealt only with images, existing methods~\cite{SORT, TAO, QDTrack} showed difficulties in classifying tracklets of rare categories in videos.
In this paper, we introduce a set classifier which re-evaluates the category of tracklet predictions given from the recent object tracker QDTrack~\cite{QDTrack}.
The plug-and-playable set classifier aggregates information from various viewpoints of an object in a similar way to \cite{SetTransformer}, which is necessary for classifying large vocabulary.
By increasing the classification accuracy of the predicted tracklets, noticeable improvements in overall accuracy are achieved in TAO and YouTube-VIS~\cite{TAO, MaskTrackRCNN}.

\section{Method}
\label{sec:method}

Our method can be easily implemented on top of existing object trackers that use two-stage detectors.
In this paper, our framework is built upon the recently proposed QDTrack~\cite{QDTrack} with integration of the additional head that can be jointly trained: the set classifier (Fig.~\ref{fig:sampling}).
The input video is first fetched by the object tracker that detects objects per-frame and generates tracklets by associating the predictions.
Next, the newly proposed set classifier receives the tracklet as a whole, where each item in a tracklet is a RoI-Aligned~\cite{MaskRCNN} feature that corresponds to a predicted box.
Finally, with the design of incorporating multiple viewpoints of instances, the set classifier head precisely predicts the category of the given tracklet.

Throughout this paper, $k^\text{th}$ regional proposal from RPN~\cite{FasterRCNN} is defined as $B_k=(b_k, c_k, i_k)$, where $b_k$ denotes the box coordinates.
Here, $c_k$ and $i_k$ denote the category and the identity of the ground-truth box that $b_k$ is matched to.


\subsection{Set Classifier}
\label{sec:set classifier}

Typical trackers~\cite{QDTrack, CenterTrack, MaskTrackRCNN}, which have been specialized to major classes (e.g., car and pedestrian), have not given much consideration to the classification.
Therefore, they usually use class predictions from detectors~\cite{FasterRCNN, CenterNet, MaskRCNN} directly.
However, failure of such naïve classification is witnessed when encountering the object tracking benchmark with long-tailed distribution~\cite{TAO}.
The most challenging aspect of classification in such a scenario is the change of appearance over time.
To overcome this problem, we propose a classifier that aggregates information of multiple views from a tracklet.
This simple extension strengthens the tracklet-level classification accuracy by mitigating the inconsistency of most trackers; deriving class prediction from naïve averaging or max-count of per-RoI classification results.

\begin{figure}[t]
\begin{center}
\includegraphics[width=\linewidth]{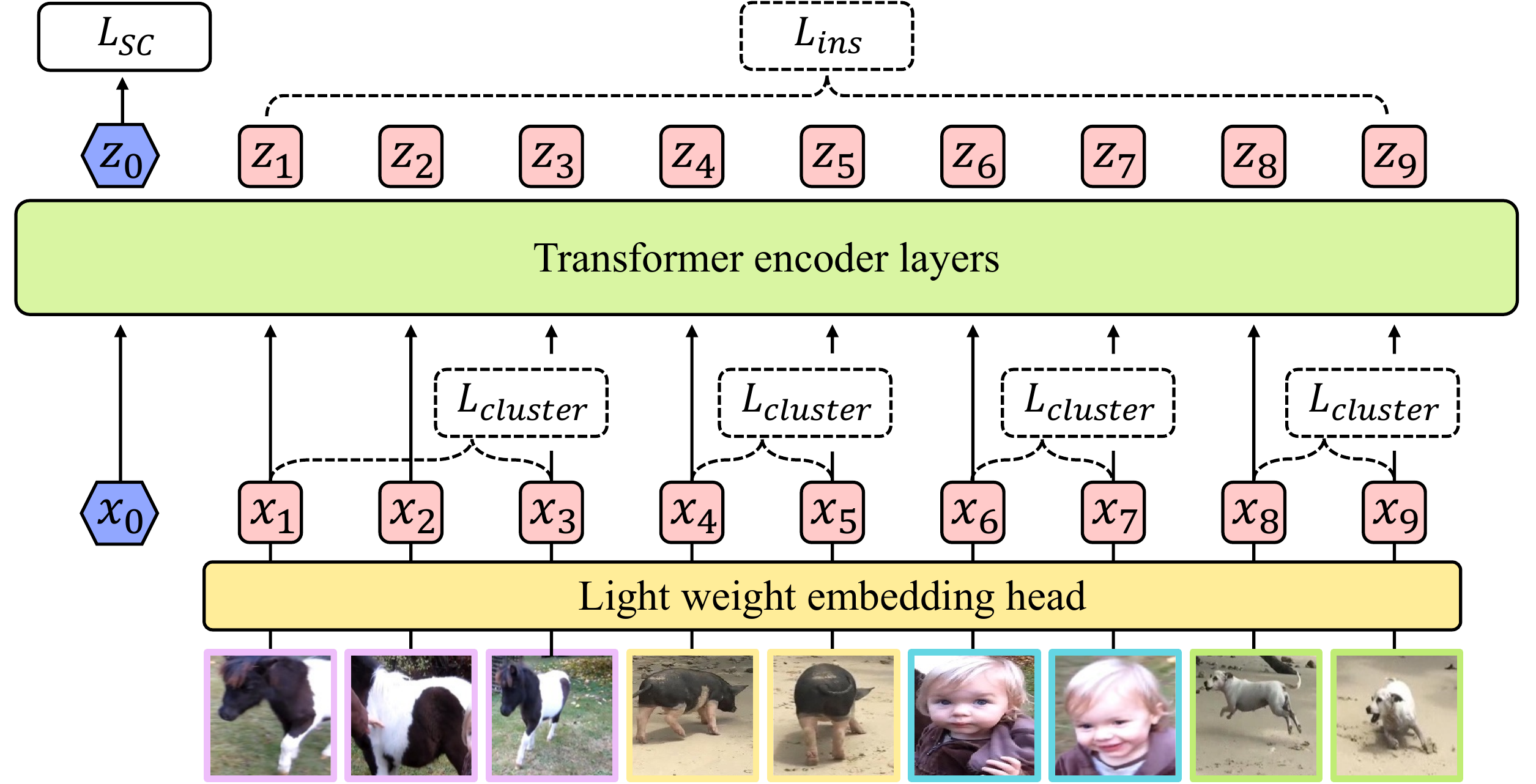}
\end{center}
\vspace{-5mm}
\caption{
The structural design of the set classifier.
$L_\textit{SC}$ is calculated from the prediction of the classification token, which embeds the overall context of a tracklet.
The use of two auxiliary losses, $L_\textit{ins}$ and $L_\textit{cluster}$, brings further improvements in accuracy.
}
\vspace{-5mm}
\label{fig:module}
\end{figure}

We design our set classifier as a stack of $N_E$ transformer encoder layers~\cite{Transformer} (\cref{fig:module}).
Inputs for the set classifier come from two sources: a classification token and RoI tokens from a tracklet.
The classification token is trainable, similar to recent uses of tokens in transformers~\cite{BERT, ViT, DeiT}.
Each RoI feature corresponding to box labels in a tracklet passes through an extra lightweight embedding head to become a RoI token.
By inserting the classification token $x_0$ and RoI tokens $\{x_l\}_{l=1}^{L}$ together into the set classifier, the classification token encodes the overall contextual information of the given tracklet and the set classifier outputs embeddings $\{z_l\}_{l=0}^{L}$.

Using the output embedding of the classification token $z_0$, logits $\hat{y} \in \mathbb{R}^{C}$ are predicted, where $C$ is the number of categories of the dataset.
With the prediction, loss can be calculated using Cross-Entropy (CE) Loss as
\begin{equation}
\label{eq:cross_entropy}
    L_\textit{SC}(y, \hat{y}) = - \sum_{c=1}^{C}y_c\log(\sigma_c),\qquad \sigma_c=\frac{e^{\hat{y}_c}}{\sum_{k=1}^{C}{e^{\hat{y}_k}}},
\end{equation}
where $y_c \in \{0, 1\}$, $1 \leq c \leq C$ is the one-hot ground truth label.

To train the set transformer, we first extract tracklets from training video clips as done in QDTrack~\cite{QDTrack}.
Then the set transformer learns to predict one most probable class for each tracklet.
This approach has several advantages over the conventional method that predicts class per frame (or RoI).
It becomes more robust against noise such as motion blur and occlusion, and accurate on classifying the tail classes by aggregating information from multiple sources.

However, due to the lack of tracklet annotations in video, using the aforementioned training pipeline alone inevitably encounters over-fitting.
Even using the largest scale long-tail video detection dataset~\cite{TAO}, we observed that the number of training tracklets is not sufficient to prevent over-fitting, especially for the tail classes.
It is unavoidable as annotating for videos requires much more effort and resources than for images.
To alleviate this issue, we propose effective data augmentation methods that can generate tracklets with video characteristics.
Our augmentation methods can not only increase the number of existing video tracklets, but also synthesize video tracklets from images.


\subsection{Tracklet Augmentations}
\label{sec:tracklet augmentations}
\begin{figure}[t]
\begin{center}
\includegraphics[width=\linewidth]{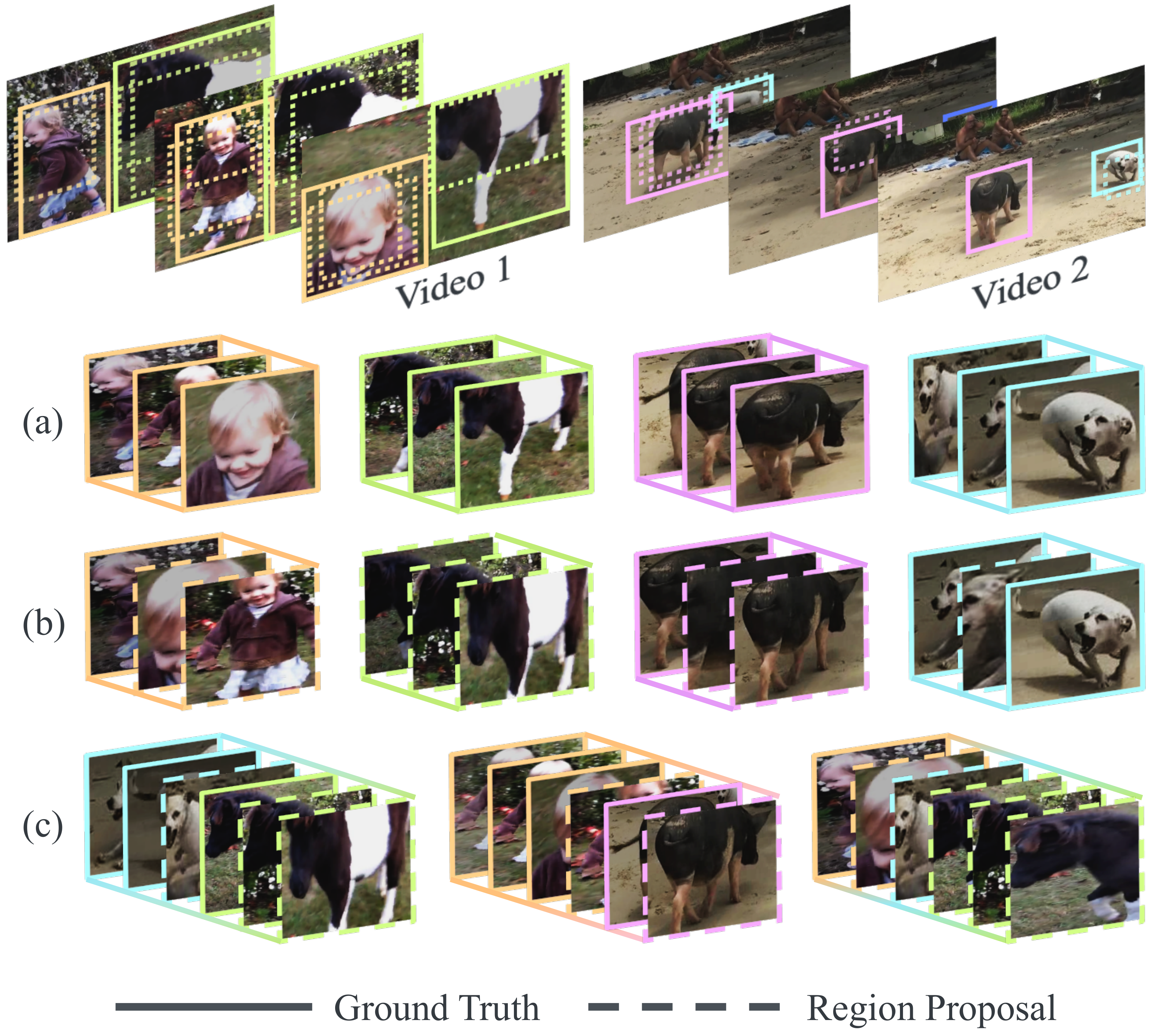}
\end{center}
\vspace{-5mm}
\caption{
Augmentations of tracklets.
The solid lines denote ground-truth box annotations, and the dotted lines are region proposals from RPN.
(a) Because annotating a video requires huge efforts, video datasets have sparse labels that are insufficient to train large vocabularies.
(b) Tracklets can be diversified from substitution of box labels with region proposals.
(c) Further supervision to tail classes can be derived from mixtures of tracklets, having multiple classes.
}
\vspace{-4mm}
\label{fig:augmentation}
\end{figure}

In this section, we describe data augmentation methods for generating tracklets to resolve the data shortage issue (\cref{fig:augmentation} (a)).
The data augmentation is designed with following considerations that are essential for effective supervision of the set classifier.

\paragraph{Tracklets from region proposals}
Instead of using only the ground-truth box labels, we combine numerous regional proposals of RPN in multiple frames to generate tracklets with new views (\cref{fig:augmentation} (b)).
Compared to sparse box annotations, numerous region proposals can be collected where each has a different perspective to the matched ground-truth box.
Therefore, this augmentation leads to quantitatively increasing and further diversifying the training samples for the set classifier.
Moreover, the augmentation is applicable not only to videos~\cite{TAO}, but also to images~\cite{LVIS}; tracklets can be generated from images.
Even if components originated from a single image, the combinations of possible object regions can imitate appearance transitions that are inherent in videos.
With the augmentation, the set classifier can learn to aggregate information from different viewpoints, which is necessary for classifying large vocabulary.

\paragraph{More diverse tracklets by mixing RoIs}
We observed that the proposed tracklet augmentation significantly impacts the classification performance.
However, it is worth noting that RPN proposes more RoIs on head classes than tail classes, which intensifies the imbalance between the head and the tail.
If tracklets are generated by gathering RoIs matched to a single identity, it inevitably leads to tail classes having fewer and less diverse tracklets than that of head classes.
To diversify the tracklets and expose more tail classes, we propose to further augment tracklets to have multi-identity and multi-class tracklets (\cref{fig:augmentation} (c)).
Inspired by augmentation techniques that mix two images~\cite{MixUp, CutMix}, we extend the idea to our augmentation. 
In other words, we do not restrict the generated tracklets to be composed of a single class.
With this approach, the number of combinations of forming tracklets is countless by mixing all the RoIs within a training batch. 
This makes our set classifier even more robust to over-fitting. 

\paragraph{Resampling for tail classes}
Our RoI mix-up method also enables unlimited resampling of sparse tail class RoIs.
Recent findings~\cite{RelayBackprop, ExploringTheLimits, LVIS, BAGS} have shown the impact of training data sampling to be critical on datasets showing long-tailed distributions.
Likewise, we observed that the strategy of sampling RoIs in generated tracklets also plays an important role.
Inspired by RFS~\cite{LVIS}, the probability of sampling an RoI is determined by the total number of training samples per category in the training set.
Let $n_c$ be the total number of training annotations in the training dataset that are labeled to category $c$.
We sample RoIs under the multinomial distribution where $p'_k$, the probability of sampling $B_k$, is defined as follows:
\begin{equation}
\label{sample_prob_eq}
    p'_k = \frac{p_k}{\sum_{j=1}^{B}{p_j}}, \qquad p_k=\sqrt{\frac{1}{n_{c_k}}}.
\end{equation}
Compared to uniformly sampling RoIs, the weighted sampling strategy results in the set classifier encountering more tracklets containing RoIs of tail classes.

\paragraph{Training with Augmented Tracklets}
Using the proposed augmentation methods, let $X = \{B_l\}_{l=1}^L$ denote a generated tracklet, where $L$ is the number of items.
As mentioned above, the set $X$ can be composed of RoIs that contain numerous different categories.
For a dataset of $C$ categories, we define the label $y$ to be the proportion of categories of items within the set $X$ as follows:
\begin{equation}
y = \left\{\dfrac{\sum_{l=1}^L\mathbbm{1}_{\{c_l=c\}}}{L}\right\}_{c=1}^{C} \in [0,1]^{C},\quad \sum_{c=1}^{C}y_c = 1.
\end{equation}
For the training with the augmented tracklets, typical CE Loss (\cref{eq:cross_entropy}) can be used with the simple replacement of the one-hot ground-truth label to the multi-class soft label.


\subsection{Auxiliary Tasks}
\label{sec:auxiliary tasks}
In addition to the soft label of augmented tracklets, we can also utilize instance-level labels that correspond to region proposals.
We further introduce two auxiliary losses, each of which improves the accuracy of the set classifier.

\paragraph{Instance-wise loss}
The structural design of the set classifier is similar to the transformer based ViT~\cite{ViT}; it is composed of a stack of transformer encoder layers and uses a classification token for the final prediction (\cref{fig:module}).
Since ViT targets the image classification task, where local patches may not hold meaningful information, only the classification token can receive losses.
Contrary to ViT, the set classifier receives tokens that are embedded from object-like regional proposals.
Therefore, other tokens, not just the classification token, can be used to calculate the instance-wise loss $L_\textit{ins}$.
Instance-wise loss adopts simple CE Loss to predict the class of the originated region, which accelerates the training of the set classifier.

\paragraph{Clustering loss}
A key challenge in training the set classifier with the augmented tracklets is that the tokens originated from various identities.
For example, in \cref{fig:module}, token $x_1$ should find relevance from $x_2$ and $x_3$ and aggregate information from the two.
However, if $x_1$ wrongly interprets that $x_4$ and $x_5$ also originated from the same instance, the accuracy of the set classifier will decrease.
In order to clarify the sources of tokens, supervising the tokens to embed object appearances can be considered following many approaches that tackle tracking tasks~\cite{QDTrack, MaskTrackRCNN}.
Nevertheless, since tracking and classification are distinct to another, use of such supervision rather harms the accuracy.

From the assumption that class distributions of RoIs that share the same origination should be similar, we lightly cluster the feature representations by utilizing the class distributions.
Using the class logits $\{\tilde{y}\}_{l=1}^L$ predicted by inserting the RoI tokens $\{x_l\}_{l=1}^L$ to a linear classifier, the features can be clustered as follows:
\begin{gather}
    L_\textit{cluster}(y_l, \tilde{y}_l) = L_\textit{CE}(y_l, \tilde{y}_l) + KL(\tilde{p}_l \Vert Q_l),\\
    Q_l = \frac{ \sum_{k=1}^{L}\mathbbm{1}_{\{i_k=i_l\}} \tilde{p}_k }{ \sum_{k=1}^{L}\mathbbm{1}_{\{i_k=i_l\}} },
\end{gather}
where $y_l$ is the ground-truth category label of $l^\text{th}$ RoI, and $\tilde{p}_l$ is a class probability distribution obtained from $\tilde{y}_l$ by the softmax function.
The use of KL divergence brings the class distributions to resemble the centroid distribution $Q$.
With this auxiliary loss, the inputs to the set classifier share similar embeddings if originated from the same object, which assists the set classifier to aggregate relevant information.

\section{Experiments}

In this section, we evaluate the proposed method using TAO~\cite{TAO} and YouTube-VIS 2019~\cite{MaskTrackRCNN}.
We demonstrate that our method achieves a huge increase in the overall performance by improving the classification performance, especially for infrequent object categories in the tail.
More experiments and details can be found in the supplementary.


\begin{table*}
\centering
{
\vspace{0mm}
\begin{tabular}{@{}l|cccccc>{\columncolor[gray]{0.9}}ccc@{}}
\toprule
Method                  & AP    & AP$_{50}$ & AP$_{75}$ & AP$_\text{s}$ & AP$_\text{m}$ & AP$_\text{l}$ & AP$_\text{r}$ & AP$_\text{c}$ & AP$_\text{f}$ \\
\midrule
QDTrack~\cite{QDTrack}  & 17.2  & 29.1      & 17.4      & 5.7           & 13.1          & 22.0          & 6.5           & 11.9          & 25.9          \\
Ours                    & 18.3  & 29.5      & 18.9      & 6.7           & 11.9          & 23.7          & 13.6          & 14.0          & 23.8          \\
\bottomrule
\end{tabular}
} 
\vspace{-3mm}
\caption{
Results of detection metrics on TAO validation set.
Following LVIS, TAO reports AP$_\text{r}$ (\textit{rare} categories with 1 to 9 videos), AP$_\text{c}$ (\textit{commmon} categories with 10 to 99 videos), and AP$_\text{f}$ (\textit{frequent} categories with $\geq$ 100 videos).
AP$_\text{s}$, AP$_\text{m}$, and AP$_\text{l}$ denote accuracies on boxes of \textit{small}, \textit{medium}, and \textit{large} sizes, respectively.
} 
\vspace{-2mm}
\label{tab:sota_det}
\end{table*}
\begin{table*}
\centering
\vspace{0mm}
{
\begin{tabular}{@{}l|ccc|ccc@{}}
\toprule
\multirow{2}{*}{Method} & \multicolumn{3}{c|}{TAO validation}                       & \multicolumn{3}{c}{TAO test}                              \\
                        & TrackAP$_{50}$    & TrackAP$_{75}$    & TrackAP$_{50:95}$ & TrackAP$_{50}$    & TrackAP$_{75}$    & TrackAP$_{50:95}$ \\
\midrule
SORT~\cite{SORT, TAO}   & 13.2              & -                 & -                 & 10.2              & 4.4               & 4.9               \\
QDTrack~\cite{QDTrack}  & 15.8              & 6.4               & 7.3               & 12.4              & 4.5               & 5.2               \\
\midrule
Ours                    & 19.9              & 8.3               & 9.6               & 15.7              & 6.8               & 7.4               \\
\bottomrule
\end{tabular}
} 
\vspace{-3mm}
\caption{
Results of tracking metrics on TAO validation and test sets. Our method outperforms previous methods by a meaningful margin.
} 
\vspace{-4mm}
\label{tab:sota_track}
\end{table*}
\begin{table}
\centering
\resizebox{\linewidth}{!}{ 
\begin{tabular}{@{}l|ccccc@{}}
\toprule
Method                                  & AP    & AP$_{50}$ & AP$_{75}$ & AR$_{1}$  & AR$_{10}$ \\
\midrule
MaskTrack R-CNN~\cite{MaskTrackRCNN}    & 31.9  & 53.7      & 32.3      & 32.5      & 37.7      \\
SipMask$^{\dagger}$~\cite{SipMask}      & 33.7  & 54.1      & 35.8      & 35.4      & 40.1      \\
CrossVIS~\cite{CrossVIS}                & 36.6  & 57.3      & 39.7      & 36.0      & 42.0      \\
VisTR~\cite{VisTR}                      & 38.6  & 61.3      & 42.3      & 37.6      & 44.2      \\
\midrule
QDTrack~\cite{QDTrack}                  & 34.4  & 55.1      & 38.4      & 33.5      & 41.6      \\
Ours                                    & 37.7  & 60.4      & 39.8      & 35.6      & 45.8      \\
\bottomrule
\end{tabular}
} 
\vspace{-3mm}
\caption{
Comparison of results on YouTube-VIS 2019 with previous methods using ResNet-101.
${\dagger}$ indicates using ResNet-50.
} 
\vspace{-3mm}
\label{tab:ytvis}
\end{table}

\subsection{Datasets}
\paragraph{TAO}
Our experiments are conducted on the long-tailed object tracking benchmark, TAO~\cite{TAO}.
TAO is a large-scale dataset that has long-tailed distributions within 482 classes that are a subset of LVIS~\cite{LVIS}.
The dataset has 500 videos, 216 classes in the training set, 988 videos, 302 classes in the validation set, and 1419 videos, 369 classes in the test set.
It is worth noting that the categories in the validation set and the test set are not subsets of the training set, so categories do not overlap.
The ability to classify such non-overlapping categories should be trained from LVIS~\cite{LVIS}.

\paragraph{YouTube-VIS 2019}
Most MOT benchmarks have a very limited number of classes: less than ten.
Therefore, we further demonstrate the importance of tracklet classification on the video instance segmentation benchmark, YouTube-VIS 2019~\cite{MaskTrackRCNN}.
The dataset is composed of 40 categories, and is composed of 2,238/302/343 videos for training/validation/test.


\subsection{Implementation Details}
All models including previous works and ours use ResNet-101~\cite{ResNet} as the backbone.
By default, the number of transformer encoder layers composing the set classifier is $N_E=3$, where each layer has 8 heads with overall depth of 512.
For the inputs of the set classifier, we first flatten region proposals and embed corresponding features by two fully connected layers~\cite{FasterRCNN}.
Thanks to the plug-and-playable design of the proposed set classifier, we attach the module on top of QDTrack~\cite{QDTrack}.
Our model is fully trained end-to-end with \cite{QDTrack} and the configurations are identical unless specified.

The training schedule of QDTrack has two phases; 24 epochs of pretraining on LVIS~\cite{LVIS}, and 12 epochs of finetuning on TAO~\cite{TAO}.
Although the set classifier can be trained only from images, the accuracy higly improves when using real video samples together (\Table{mix_training}).
Therefore, while following the training schedule of~\cite{QDTrack}, we simultaneously insert videos of TAO during the pretraining phase.
For each iteration of training, the tracklet generator produces 256 augmented tracklets of varying length from 16 to 32.
The weight of $L_\textit{SC}$, $L_\textit{ins}$, and $L_\textit{cluster}$ are 0.05, 0.02, 0.1 respectively.
During inference, we follow the pipeline of generating tracklets and re-classifying the category as mentioned in ~\cref{sec:method}.
Denoting the predicted classification score from the set classifier as $c$, and the confidence score of the tracklet from \cite{QDTrack} as $s$, the output score can be obtained as $c^{\lambda_c}s^{\lambda_s}$, where $\lambda_c=\frac{1}{3}, \lambda_s=\frac{2}{3}$ by default.
We finalize the output score by multiplying the length of each to penalize redundant short tracklets.


\subsection{Main Results}
As mentioned in \cref{sec:method}, we receive tracklet predictions from the tracker, and assign a newly predicted category from the set classifier to each box that forms tracklets.
Therefore, the only difference between the outputs of QDTrack~\cite{QDTrack} and ours is the classification labels.
Surprisingly, with the simple re-classification of tracklets, our method highly improves the overall accuracy of QDTrack on both TAO and YouTube-VIS benchmarks.

\paragraph{TAO}
The results on TAO can be largely divided into two as shown in \Table{sota_det} and \Table{sota_track}, where each refers to the accuracies on detection and tracking, respectively.
Since our method is built on top of QDTrack~\cite{QDTrack}, the quality of predicted box coordinates is similar to that of~\cite{QDTrack}.
However, from the reclassification of the boxes with the set classifier, our model outperforms QDTrack with a noticeable gap of 1.1\% detection AP as shown in \Table{sota_det}.
More importantly, the score of AP$_r$ is more than doubled from that of QDTrack.
AP$_r$ is an important criterion in benchmarks of long-tailed distributions~\cite{LVIS, TAO} as it represents the ability of classifying rare categories.
From the noticeably enhanced classification results, our model achieves the best accuracy of 19.9\% on the tracking metric TrackAP$_{50}$, which is a 4.1\% of improvement compared to QDTrack.
These results signify that from the utilization of multiple information in tracklets, the set classifier successfully classifies large vocabularies.


\paragraph{YouTube-VIS 2019}
The results on YouTube-VIS 2019 benchmark is shown in \Table{ytvis}.
Since QDTrack~\cite{QDTrack} is an object tracker, it is not capable of generating segmentation masks by default.
Therefore, we attach the simple mask head of Mask R-CNN~\cite{MaskRCNN} on top of QDTrack.
From the small modification, QDTrack shows the accuracy of 34.4\% AP, which is comparable to previous methods that mainly target the video instance segmentation task.

We also demonstrate the effectiveness of our method, re-classifying tracklets using the set classifier, on the VIS benchmark.
Because the VIS dataset has relatively small vocabulary size (40 categories), an insignificant impact is expected from the use of the set classifier.
Nevertheless, the predicted class labels from our set classifier greatly improves the accuracy of QDTrack by 3.3\% AP.
The achieved score of 37.7\% AP is higher than many existing VIS models~\cite{MaskTrackRCNN, SipMask, CrossVIS}, and also competitive to recently proposed VisTR~\cite{VisTR}.


\subsection{Ablation Studies}

Using the validation set of TAO, we show various ablation studies on the set classifier and the tracklet generator.
We show how the accuracy of the set classifier can be improved by differentiating tracklet augmentations and training configurations.

\paragraph{Multi-class tracklet generation}
Here, we study the impact of the multi-class tracklet augmentation (\Table{single_multi}).
Due to the sparse number of annotated boxes, using only ground-truth box labels directly leads to severe overfitting.
Therefore, we proposed various augmentations that can enormously diversify the tracklets in \cref{sec:tracklet augmentations}.
The augmentation of tracklets using regional proposals, but not allowing them to be composed of multiple identities and multiple classes (\cref{fig:augmentation} (b)) leads to the accuracy of 17.7\% TrackAP$_{50}$.

Mixing RoIs of different identities (\cref{fig:augmentation} (c)) can be divided into two, depending on whether composing tracklets with RoIs of multiple classes is granted or not.
As shown in~\Table{single_multi}, we find the allowance of multiple classes improves the accuracy by 2.2\% TrackAP$_{50}$, while the preclusion of multiple classes rather decreases TrackAP$_{50}$.
It is because the composition of multiple classes is an important factor to diversify augmented tracklets, especially when dealing with large vocabulary scenarios where only a few regional proposals can be gathered in the tail.

\begin{table}
\centering
\resizebox{\linewidth}{!}{ 
\begin{tabular}{@{}cc|ccc@{}}
\toprule
Multi-Identity  & Multi-Class   & TrackAP$_{50}$    & TrackAP$_{75}$    & TrackAP$_{50:95}$ \\
\midrule
            &                   & 17.7              & 7.4               & 8.5\\ 
\checkmark  &                   & 17.2              & 7.9               & 8.5\\ 
\checkmark  & \checkmark        & \textbf{19.9}     & \textbf{8.3}      & \textbf{9.6}\\
\bottomrule
\end{tabular}
} 
\vspace{-3mm}
\caption{
Comparison of different tracklet augmentations.
\textit{Multi-Identity} denotes that tracklets can be composed of regional proposals from different identities.
\textit{Multi-Class} denotes that tracklets can be mixtures of multiple classes.
} 
\vspace{-1mm}
\label{tab:single_multi}
\end{table}
\begin{table}
\centering
\resizebox{\linewidth}{!}{ 
\begin{tabular}{@{}c|ccc@{}}
\toprule
\#tracklets / batch  & TrackAP$_{50}$    & TrackAP$_{75}$    & TrackAP$_{50:95}$ \\
\midrule
32              & 17.7              & 8.2               & 9.1\\
64              & 18.3              & 7.6               & 8.9\\
128             & 18.5              & 8.2               & 8.9\\
256             & \textbf{19.9}     & \textbf{8.3}      & \textbf{9.6}\\
\bottomrule
\end{tabular}
} 
\vspace{-3mm}
\caption{
Comparison of number of generated tracklets per batch.
} 
\vspace{-3mm}
\label{tab:num_sets}
\end{table}

\paragraph{Number of augmented tracklets}
Thanks to the multi-class tracklet augmentation, we can now generate training samples for the set classifier without limitations.
Therefore, the number of tracklets can be as many as the set classifier requires for training.
As shown in~\Table{num_sets}, by increasing the number from 32 to 256, TrackAP$_{50}$ increases by 2.2.
This experiment indicates that training the set classifier with more number of tracklets brings further supervision, leading to higher accuracies.

\paragraph{Importance of video training}
The main purpose of the set classifier is to aggregate information from multiple viewpoints of an object.
Though the ability can be partially gained from augmented tracklets that are generated from images, such tracklets cannot serve the true appearance changes in videos.
However, as mentioned earlier, using only the sparse labels from videos leads to an overfitting (6.5\% TrackAP$_{50}$).
Therefore, we supply videos of TAO to generate more meaningful tracklets during the pretraining phase.
As shown in \Table{mix_training}, a noticeable improvement of 2.8\% TrackAP$_{50}$ is gained by using the augmented tracklets from videos.

\begin{table}
\centering
\resizebox{\linewidth}{!}{ 
\vspace{0mm}
\begin{tabular}{@{}cc|ccc@{}}
\toprule
Images      & Videos        & TrackAP$_{50}$    & TrackAP$_{75}$    & TrackAP$_{50:95}$ \\
\midrule
\checkmark  &               & 17.1              & 6.9               & 8.4               \\
            & \checkmark    & 6.5               & 2.6               & 2.8               \\
\checkmark  & \checkmark    & \textbf{19.9}     & \textbf{8.3}      & \textbf{9.6}      \\
\bottomrule
\end{tabular}
} 
\vspace{-3mm}
\caption{
Comparison of different domains used for training the set classifier.
\textit{Images} and \textit{Videos} denote the use of LVIS~\cite{LVIS} and TAO~\cite{TAO}, respectively.
} 
\vspace{-1mm}
\label{tab:mix_training}
\end{table}
\begin{table}
\centering
\resizebox{\linewidth}{!} 
{
\begin{tabular}{@{}c|ccc@{}}
\toprule
Tracklet Length             & TrackAP$_{50}$    & TrackAP$_{75}$    & TrackAP$_{50:95}$ \\
\midrule
$\left[\:\:8, 16\right)$    & 19.0              & 7.8               & 8.7\\
$\left[16, 32\right)$       & \textbf{19.9}     & \textbf{8.3}      & \textbf{9.6}\\
$\left[32, 64\right)$       & 18.3              & 7.0               & 8.3\\
\bottomrule
\end{tabular}
} 
\vspace{-3mm}
\caption{
Comparison of different lengths of generated tracklets.
} 
\vspace{-3mm}
\label{tab:seq_length}
\end{table}

\paragraph{Length of augmented tracklets}
In~\Table{seq_length}, we investigate how the length of augmented tracklets used during training affect the final accuracy.
In order to robustly classify the outputs of the tracker with varying lengths, we provide augmented tracklets of different lengths during training.
On the TAO validation dataset, the total average length of tracklets predicted by QDTrack is 21.24.
As can be expected from the average length, using tracklets ranging from 16 to 32 for training turns out to bring the best performance.

\paragraph{Sampling ratio of tail categories}
Sampling images containing tail classes more frequently  similar to RFS~\cite{LVIS}, we regulate the generation of tracklets to favor including tail classes over that of head classes (\cref{sec:tracklet augmentations}).
The regulation can be controlled by differentiating the probability of sampling RoIs that are used in the tracklet augmentation.
The probability is defined using the total number of training annotations per class $n_c$ as denoted in~\cref{sample_prob_eq}.
The results of controlling the probability is shown in~\Table{sample_prob}.
If $p=0$, all RoIs are given the same probability to be sampled, which indicates uniform sampling.
Since uniform sampling does not consider a huge imbalance of annotations between different categories, it shows a large reduction of accuracy.
Among many $p$ values, 0.5 turns out to be the best hyperparameter that well alleviates the imbalance and achieves the highest performance.

\paragraph{Use of auxiliary losses}
We study the effects of the introduced auxiliary losses: $L_\textit{ins}$ and $L_\textit{cluster}$.
As described in~\cref{sec:auxiliary tasks}, the two losses assist the training of the set classifier.
From the baseline which does not use the two losses, the use of $L_\textit{ins}$ brings an increment of 0.2\% TrackAP$_{50}$.
Further improvement of 1.0\% TrackAP$_{50}$ is gained by additionally adopting $L_\textit{ins}$, which assists the set classifier to easily aggregate relevant information.

\begin{table}
\centering
\vspace{0mm}
{ 
\begin{tabular}{@{}c|ccc@{}}
\toprule
$p$             & TrackAP$_{50}$    & TrackAP$_{75}$    & TrackAP$_{50:95}$ \\
\midrule
uniform         & 15.6              & 7.3               & 7.8\\ 
$0.25$          & 17.3              & 7.2               & 8.4\\ 
$0.5$           & \textbf{19.9}     & \textbf{8.3}      & \textbf{9.6}\\ 
$0.75$          & 17.8              & 7.1               & 8.3\\ 
$1.0$           & 17.5              & 7.5               & 8.5\\ 
\bottomrule
\end{tabular}
} 
\vspace{-3mm}
\caption{
Comparison of assigning different probabilities used by the multinomial sampling; $n_{c}^{-p}$ in~\cref{sec:tracklet augmentations}.
As $p$ increases, the sampling policy favors RoIs of tail classes.
} 
\vspace{-1mm}
\label{tab:sample_prob}
\end{table}
\begin{table}
\centering
\resizebox{\linewidth}{!}{ 
\begin{tabular}{@{}cc|ccc@{}}
\toprule
Instance     & Cluster           & TrackAP$_{50}$    & TrackAP$_{75}$    & TrackAP$_{50:95}$ \\
\midrule
             &                   & 18.7              & 7.8               & 8.8\\ 
\checkmark   &                   & 18.9              & 8.0               & 9.0\\
\checkmark   & \checkmark        & \textbf{19.9}     & \textbf{8.3}      & \textbf{9.6}\\ 
\bottomrule
\end{tabular}
} 
\vspace{-3mm}
\caption{
Comparison of using the auxiliary losses.
\textit{Instance} and \textit{Cluster} denote the use of $L_\textit{ins}$ and $L_\textit{cluster}$, respectively.
} 
\vspace{-3mm}
\label{tab:aux_loss}
\end{table}

\section{Conclusions}

In this paper, we demonstrated that classification is a key factor in tracking performance on benchmarks with large vocabulary, and introduced the set classifier that takes the whole spatio-temporal features of a tracklet.
The set classifier precisely classifies large vocabularies by aggregating information from multiple viewpoints.
In order to bring out the potential of the set classifier, we also proposed tracklet augmentations that greatly diversify sparse annotations.
Furthermore, we strengthen the supervision with the suggested auxiliary losses.
The set classifier achieves the new state-of-the-art accuracy on the challenging benchmark TAO, and also shows competitive results on YouTube-VIS 2019.
For future work, we plan to design a video-targeted classifier that can precisely classify large vocabulary while being capable of online inferencing.

\section*{Acknowledgements}
This work was partly supported by Institute of Information \& communications Technology Planning \& Evaluation (IITP) grant funded by the Korea government(MSIT), Artificial Intelligence Innovation Hub under Grant 2021-0-02068, Artificial Intelligence Graduate School Program under Grant 2020-0-01361, and Development of High Performance Visual BigData Discovery Platform for Large-Scale Realtime Data Analysis under Grant 2014-3-00123.
{
    \clearpage
    \small
    \bibliographystyle{ieee_fullname}
    \bibliography{macros,reference}
}
\newpage



\end{document}